\documentclass[runningheads]{llncs}
\usepackage[T1]{fontenc}
\usepackage{graphicx}
\usepackage[table, xcdraw, pdftex, usenames, dvipsnames]{xcolor}
\usepackage[nocompress]{cite}
\usepackage{amsmath,amsfonts}
\usepackage{subcaption}
\usepackage{booktabs}
\usepackage{multirow}

\begin{document}
\title{Run-Time Monitoring of ERTMS/ETCS\\Control Flow by Process Mining}
\author{Francesco Vitale\inst{1}\orcidID{0000-0003-2325-0056} \and
Tommaso Zoppi\inst{2}\orcidID{0000-0001-9820-6047} \and
Francesco Flammini\inst{2,3}\orcidID{0000-0002-2833-7196}\and
Nicola Mazzocca\inst{1}\orcidID{0000-0002-0401-9687}}

\authorrunning{F. Vitale et al.}

\institute{University of Naples Federico II, Naples, Italy\\\email{\{name.surname\}@unina.it} 
\and
University of Florence, Florence, Italy\\\email{\{name.surname\}@unifi.it}
\and
University of Applied Sciences and Arts of Southern Switzerland, Lugano, Switzerland\\\email{francesco.flammini@supsi.ch}}
\maketitle

\begin{abstract}
Ensuring the resilience of computer-based railways is increasingly crucial to account for uncertainties and changes due to the growing complexity and criticality of these systems. Although their software relies on strict verification and validation processes following well-established best-practices and certification standards, anomalies can still occur at run-time due to residual faults, system and environmental modifications that were unknown at design-time, or other emergent cyber-threat scenarios. This paper explores run-time control-flow anomaly detection using process mining to enhance the resilience of ERTMS/ETCS L2 (European Rail Traffic Management System / European Train Control System Level 2). Process mining allows learning the actual control flow of the system from its execution traces, thus enabling run-time monitoring through online conformance checking. In addition, anomaly localization is performed through unsupervised machine learning to link relevant deviations to critical system components. We test our approach on a reference ERTMS/ETCS L2 scenario, namely the RBC/RBC Handover, to show its capability to detect and localize anomalies with high accuracy, efficiency, and explainability.


\end{abstract}

\section{Introduction}
\label{sec:intro}
The European Rail Traffic Management System (ERTMS)\footnote{\url{https://www.era.europa.eu/domains/infrastructure/european-rail-traffic-management-system-ertms\_en}} is a European standard setting the specification of both architecture and functions for interoperable, efficient, and dependable railways. The European Train Control System (ETCS) standardizes the Automatic Train Protection (ATP) subsystem \cite{ghazel2014formalizingertmsetcsspecs}. ERTMS/ETCS has different levels of operation, which determine how the on-board equipment of trains exchanges key information with the trackside subsystems. In this paper, we address ERTMS/ETCS L2, which uses wireless communication between on-board equipment and the Radio Block Center (RBC), based on the Euroradio protocol, to ensure continuous signalling. Train position is determined through Eurobalises and sent to the RBC via the Position Reports (PRs); RBC computes and transmits to the trains the so-called Movement Authority (MA), when full supervision is active (i.e., after start of mission and in case of no failures).

ERTMS/ETCS L2 must undergo strict verification and validation processes according to well-established best-practices, internal standards, and certification requirements. These include extensive requirements engineering, modeling and formal verification also using model-checking \cite{angeletti2010safetycriticalsoftwarecoverageanalysis, cimatti2012ertmsl2verificationvalidation, flammini2014multiformalismapproach, ghazel2014formalizingertmsetcsspecs, karg2016modeldrivenseopenetcs, campanile2022textbasedmetrics}, as well as simulation and testing \cite{ditommaso2005anomaliessimulation, valdivia2017etcssafetytesting, ameurbolifa2019verifyingetcssoftware, gaspari2019hybrideuropeanrailtrafficmgms, nardone2020systemlevelfunctionaltesting, su2022safetyinterlocking}. 
However, the coverage of these activities can never be complete due to several factors, such as: (1) the usage of natural language specification that might have ambiguities; (2) the high level of complexity undermining full formal verification and testing coverage, (3) heterogeneity in components developed by different manufacturers using diverse software versions and implementations. In such a scenario, residual or interaction faults, unexpected or uncontrolled modifications, system or environmental uncertainties, different specification interpretation or implementation across multiple manufacturers, emergent cyber-threats, as well as any other ``unknown unknowns'', pose threats that might affect the correct execution of ERTMS/ETCS L2 procedures and generate failures. Generally speaking, functional, environmental and technological changes during system operation require appropriate means to ensure resilience, i.e., the persistence of dependability when facing changes \cite{laprie2005resilience}.

To address the problem of ensuring dependability in the presence of unexpected events possibly corrupting ERTMS/ETCS L2 control flow, we propose an approach for run-time monitoring and anomaly detection through process mining, which encompasses a set of explainable algorithms to model the so-called normative behavior (i.e., process discovery), and to check any deviations during system operation (i.e., conformance checking) \cite{aalst2022pmhb}. In fact, process mining allows:
\begin{itemize}
    \item Modeling the execution of ERTMS/ETCS L2 procedures under normal conditions through process discovery;
    \item Collecting run-time ERTMS/ETCS L2 control-flow and checking execution correctness against the normative execution model through conformance checking.
\end{itemize}

The characterization of ERTMS/ETCS L2 procedures under normal conditions from actual execution traces helps bridge the gap between specifications/models and actual system implementations. Although model-driven engineering is a widely adopted practice when developing safety-critical systems, ensuring strict consistency between system models and their software implementations is challenging due to the system's complexity and the concurrent operation of multiple developers \cite{weidmann2018tolerant,trols2019multifaceted}. As a result, some degree of divergence between specifications/models and the actual software is often unavoidable. This makes the characterization of the system’s control flow from execution traces particularly valuable. In addition, while traditional rule-based methods (e.g., decision trees) provide a static representation of if--else rules \cite{witten2016mltools} and more complex machine learning methods (e.g., recurrent neural networks) either lack an explicit representation of control-flow patterns or the ability to provide a process-based, trustworthy explanation of the deviations from prescriptive behavior \cite{rawal2022trustworthyai,aalst2022pmhb}, process mining explicitly allows capturing behavioral patterns --- such as concurrent and exclusive control flow --- through process discovery algorithms. Thereby, process discovery allows building prescriptive process models using well-known and interpretable formalisms, such as the Petri net. Finally, the approach enables resilience to changes, uncertainties, and any ``unknown unknowns'' as it combines data-driven insights with model-based analyses, taking concrete steps toward the development of an explainable run-time monitor. More specifically, conformance checking algorithms can pinpoint local diagnoses when comparing run-time control-flow with the prescriptive process model built with process discovery, supporting the root cause analysis of abnormal behavior.

Process mining has been applied for anomaly detection in several domains, including business processes of organizations \cite{bezerra2013adlogspais, vitale2025cfad}, software applications \cite{cinque2019hiddenapplicationerrorspm, pecchia2020applicationfailuresanalysispm}, computer networks \cite{saintpierre2014dnspmad, hemmer2021adpm}, and cyber-physical systems \cite{myers2018icsadpm, vitale2025pmdt}. Notably, process mining has been also applied using an ERTMS/ETCS dataset generated from a high-level description of one use-case scenario in \cite{debenedictis2023dtadiiot, vitale2025cfad}. 
However, previous work did not thoroughly analyze ERTMS/ETCS L2 use cases or investigate process mining support for their development, monitoring, and the explanation of control-flow deviations from prescriptive specifications. In contrast, we propose a dedicated methodology that systematically combines behavior characterization, run-time monitoring, and anomaly detection and explanation. Our approach leverages offline ERTMS/ETCS L2 simulation or execution to extract a high-fidelity execution model of the target use case, integrates run-time execution with online conformance checking, applies machine learning-based post-processing of conformance diagnoses using clustering algorithms, and ultimately labels and localizes anomalies to pinpoint the faulty ERTMS/ETCS L2 components.

This paper is structured as follows. Section \ref{sec:background} establishes the background on key dependability concepts, ERTMS/ETCS L2, machine learning and process mining; Section \ref{sec:methodology} describes the proposed methodology for run-time monitoring of ERTMS/ETCS control flow; Section \ref{sec:case_study} shows the application of the methodology to RBC/RBC Handover, the chosen ERTMS/ETCS L2 scenario; Section \ref{sec:experimentation} evaluates the anomaly detection and localization capabilities of the proposal; and Section \ref{sec:conclusions} draws the conclusions and presents future work.

\section{Background and Motivation}
\label{sec:background}
This section establishes the preliminary definitions used throughout the paper, and background information on ERTMS/ETCS L2, process mining, and machine learning. 

\subsection{Preliminary Definitions}
Traditional definitions of \textbf{dependability} typically emphasize either “the ability to deliver service that can justifiably be trusted” or “the ability to avoid service failures that are more frequent and more severe than is acceptable” \cite{avizienis2004dependabilitytaxonomy}. These definitions encompass a broad range of strategies for achieving dependability, including fault prevention, fault tolerance, fault removal, and fault forecasting. However, they do not explicitly address a key concept that is central to this paper: resilience. \textbf{Resilience} refers to a system’s ability to continue delivering correct service in the face of changes in its run-time environment, including functional, environmental, and technological changes \cite{laprie2005resilience}.

To achieve resilience in modern computer-based railways, including those based on ERTMS/ETCS L2, we propose to add specific software monitors performing \textbf{run-time control-flow anomaly detection}. Anomaly detection is a broad research area that aims to find ``patterns in data that do not conform to a well defined notion of normal behavior" \cite{chandola2009ad}. Different types of anomaly detection address different types of anomalies. In this paper, we address control-flow anomalies, namely those anomalies that manifest as unknown, skipped, and wrongly-ordered activities of event logs from computer-based systems \cite{vitale2025cfad}. In addition, we aim to perform control-flow anomaly detection at run-time, including when the system is operating in non perfectly known/predictable conditions, hence subject to functional, environmental, and technological uncertainties. Finally, it is worth mentioning that we propose an \textbf{unsupervised} approach to anomaly detection and localization, as we aim to characterize normal behavior from reference execution traces of the target system, and localize anomalies based on the model-based diagnoses we perform against a reference process model of the software.

\subsection{ERTMS/ETCS L2}
ERTMS/ETCS is characterized by a certain level of operation, from 1 to 3, which defines the degree of automation performance, and the required trackside and on-board equipment \cite{ghazel2014formalizingertmsetcsspecs}. 
\begin{figure*}[!t]
\centering
\includegraphics[width=0.6\textwidth]{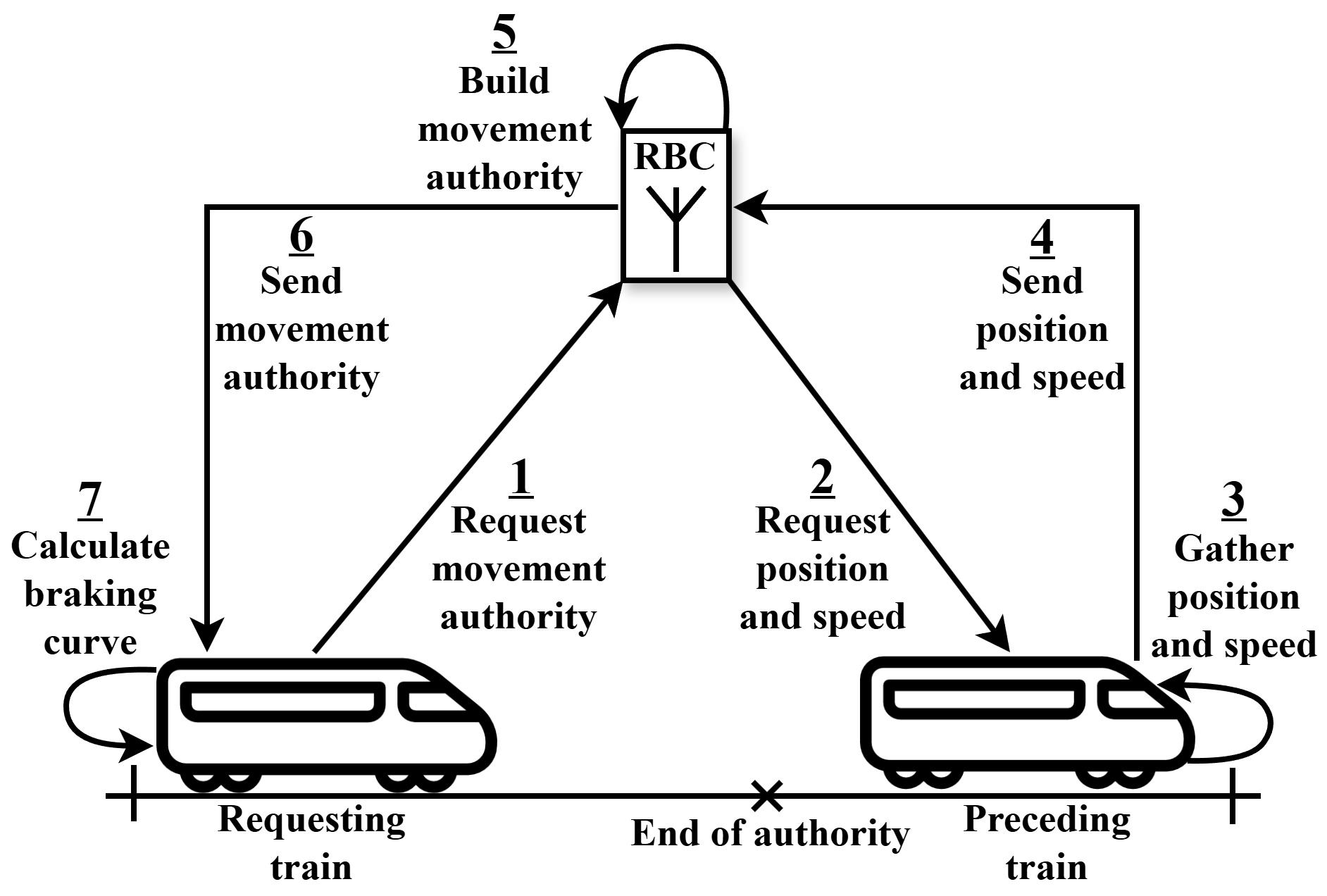}
\caption{Movement authority provision in ERTMS/ETCS L2.}
\label{ERTMS_MA_PROVISION}
\end{figure*}
In this paper, we focus on the most successful level of operation to date, which is level 2. At this level, the on-board system collects the train's position through Eurobalises, which are electronic beacons installed across the tracks. The PR is forwarded to the RBC, which is responsible for train separation; the RBC elaborates the MA based on track occupation. Fig. \ref{ERTMS_MA_PROVISION} shows a common use-case in ERTMS/ETCS, namely the MA request and provision to supervise train traffic. This use-case can be implemented within the Start-of-Mission procedure, which is one of the procedures documented in specification SUBSET-026; such a subset prescribes the behavior of ERTMS/ETCS constituents in all reference operational scenarios. Each step, labeled 1 to 7, requires either exchanging information between the on-board equipment of the requesting and preceding trains with the RBC, or using the on-board modules to perform some kind of data processing. The most important task performed on-board is the calculation of the so-called braking curve (also known as dynamic speed profile).

We assume that several types of run-time anomalies may occur across the many modules that compose ERTMS/ETCS L2, such as unexpected train data exchanged between the Driver Machine Interface (DMI) and the RBC through the Radio Transmission Module (RTM), mispositioning of balises,
and issues in establishing connections with the RBC \cite{ditommaso2005anomaliessimulation, valdivia2017etcssafetytesting, cai2019reliabilityavailabilityeval}. These anomalies may propagate across different components, and may influence other key modules, such as the European Vital Computer (EVC), which is the on-board component responsible for performing critical operations, such as calculating the braking curve.

Among the ERTMS/ETCS L2 operational scenarios of interest, we consider the RBC/RBC Handover, which is critical because it transfers train supervision to the ``accepting'' RBC from the ``handing-over'' RBC, when the train leaves the area covered by the handing-over RBC. The procedure includes: pre-announcement of the transition by the handing-over RBC; registration to the new communication network; generation of movement authorities; announcement of the RBC transition; transfer of train supervision to the accepting RBC; and termination of the session with the handing-over RBC. There are several criticalities linked to the RBC/RBC handover use case:

\begin{itemize}
    \item Different implementations of software modules developed by independent vendors;
    \item Software defects and unverified faulty edge-cases;
    \item Unpredictable environmental changes.
\end{itemize}

These aspects introduce uncertainties that cannot be completely avoided despite of strict model-checking and extensive testing activities. 

\subsection{Process Mining and Machine Learning}
Process mining ``aims to improve operational processes through the systematic use of event data'' by mainly two types of algorithms: process discovery and conformance checking \cite{aalst2022pmhb}. Process discovery attempts to build a normative behavioral model from historical event data collected by monitoring a reference application or process. Conformance checking aligns new event data to the normative model to diagnose any deviations from the control flow captured by the model. In addition, conformance checking can be performed online, i.e., it can be applied to streaming event data through, e.g., window-based approaches and prefix alignments \cite{burattin2020streamingpdcc, vanzelst2019onlinecc}.

Machine learning allows encoding data patterns into models. Traditional machine learning approaches capture patterns in data through rule-based learning, probabilistic modeling, linear and non-linear modeling, and clustering \cite{witten2016mltools}. Advanced techniques include deep learning, which implements (deep) artificial neural networks mimicking the functioning of the human brain \cite{dong2021dlsurvey}. Machine learning algorithms also differ based on their learning paradigm (supervised, unsupervised, and semi-supervised) and degree of explainability \cite{rawal2022trustworthyai}. 

Previous works have shown that conformance checking can be combined with unsupervised machine learning for explainable control-flow anomaly detection. The approach proposed in this paper is inspired by the work reported in reference \cite{vitale2025cfad}, and proposes performing unsupervised run-time monitoring of ERTMS/ETCS L2 control flow through the combination of conformance checking and clustering.

\section{Methodology}
\label{sec:methodology}
\begin{figure*}[!t]
\centering
\includegraphics[width=\textwidth]{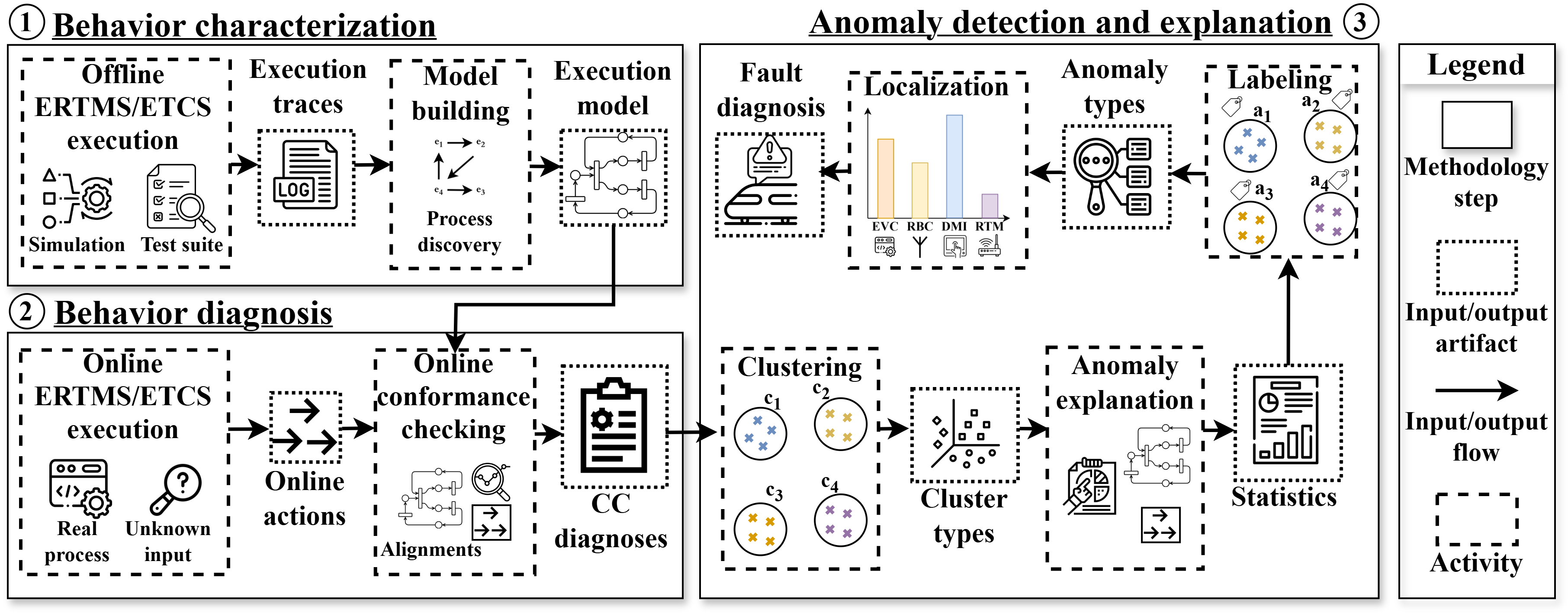}
\caption{The proposed methodology for run-time monitoring of ERTMS/ETCS control flow by process mining.}
\label{METHODOLOGY}
\end{figure*}

The methodology proposed in this paper aims to ultimately provide fault diagnosis at run-time through an explainable approach based on process mining and clustering. The methodology is organized into three steps, namely behavior characterization, behavior diagnosis, and anomaly detection and explanation.

\subsection{Behavior Characterization}
\subsubsection{Offline ERTMS/ETCS Execution} The first step begins with offline ERTMS/ETCS execution, where a test suite is applied to a simulation environment or an actual system implementation. The simulation of ERTMS/ETCS systems can provide different types of high-level events that can be used to characterize the execution model of the target scenario. For example, di Tommaso et al. \cite{ditommaso2005anomaliessimulation} simulated the Track Ahead Free (TAF) procedure with a trackside simulation environment. In this procedure, the RBC sends a TAF request to the train driver through the DMI to possibly grant a full supervision MA. Gaspari et al. \cite{gaspari2019hybrideuropeanrailtrafficmgms} exemplified another remarkable simulation approach, involving the execution of abstract state machines representing several ERTMS/ETCS L3 scenarios to test the correctness of their models. The execution traces are collected to model the software behavior. We define a set of $k$ traces as $\Sigma = \{\sigma_1, \sigma_2, \dots, \sigma_k\}$, where each trace $\sigma$ is an ordered sequence of events: $\sigma = \langle e_1, e_2, \dots, e_{|\sigma|} \rangle \in \Sigma$, with $e_j$ representing the $j$-th event. The definition of an event depends on the level of abstraction used in simulating ERTMS/ETCS L2. Since these systems involve distributed software components communicating over a railway network, events may correspond to specific software procedures executed by these components. This is a realistic representation, as safety-critical ETCS software is usually implemented in C/C++ \cite{angeletti2010safetycriticalsoftwarecoverageanalysis, cimatti2012ertmsl2verificationvalidation, karg2016modeldrivenseopenetcs}.

\subsubsection{Model Building} After the collection of execution traces, the model building activity applies a process discovery algorithm $\gamma$ to find an execution model $M$ from the set of traces representing the actual behavior of the system, i.e., $\gamma(\Sigma)=M$. There are many process discovery algorithms classified by their different learning approaches, including footprint-based and directly-follows-graph-based approaches that uncover local relationships between events (e.g., the $\alpha$-miner and split miner), divide-and-conquer approaches that iteratively find relationships starting from the most general one (e.g., the inductive miner and partially ordered workflow language miner), and region-based approaches that obtain so-called regions from either state-based or language-based represesentations of the input log (e.g., the integer linear programming-based miner) \cite{aalst2022pmhb}. Each process discovery algorithm has its own representational bias, which influences the quality of the resulting execution model. The execution model can be, e.g., a Petri net, a Business Process Modeling Notation (BPMN) model, or a process tree. The target formalism depends on the requirements of the applications. For example, a Petri net may be better suited for safety-critical environments due to its formal semantics. Hence, we consider the Petri net as the target formalism in the following.

\subsection{Behavior Diagnosis}
\begin{figure*}[!t]
\centering
\includegraphics[width=0.85\textwidth]{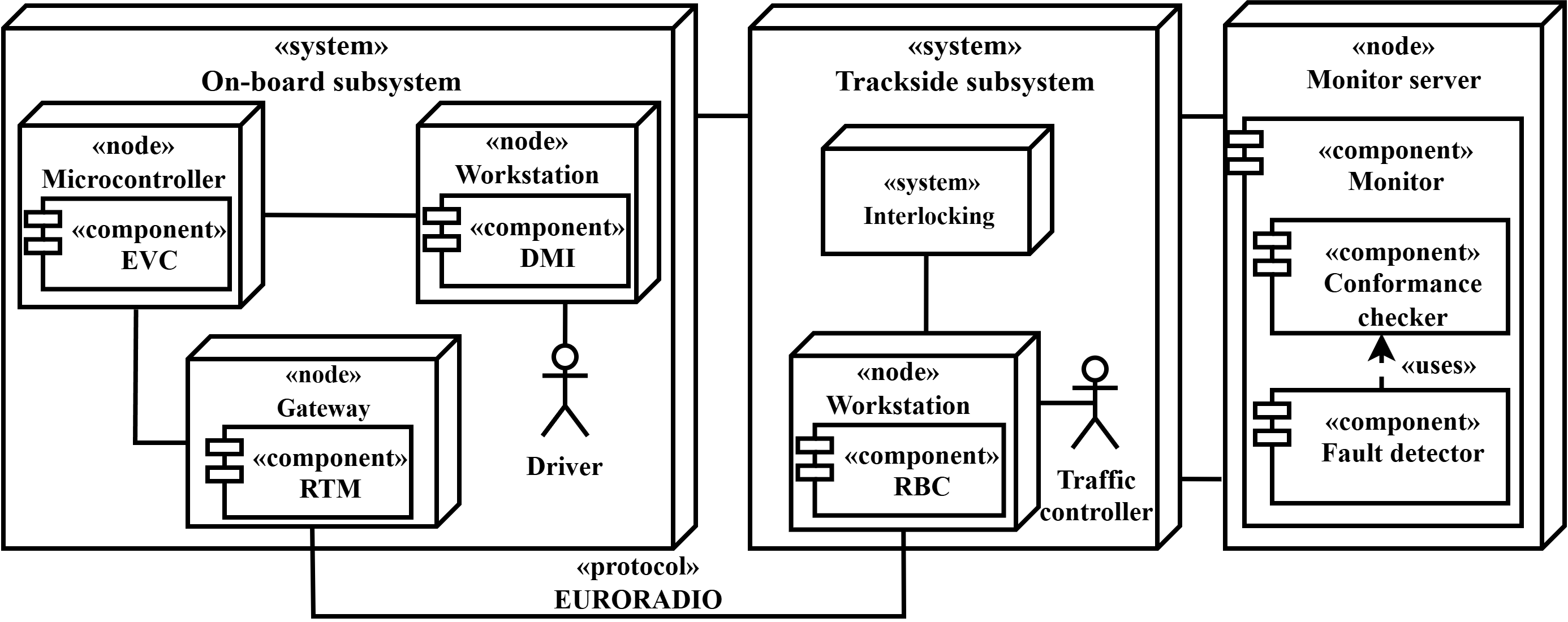}
\caption{An example deployment diagram showing the on-board and trackside subsystems communicating with each other and sending run-time events to a monitor server.}
\label{ERTMS_DEPLOYMENT_DIAGRAM}
\end{figure*}

\subsubsection{Online ERTMS/ETCS Execution}The second step is initiated by the online execution of an ERTMS/ETCS procedure with an unknown input. The real process is monitored throughout its execution. Such run-time monitoring can be performed in different ways. A plausible mechanism would be source-code instrumentation by well-defined logging rules, as proposed in \cite{cinque2013rulebasedmonitoring}. This technique places logging rules to record specific events occurring in a program, such as interactions between objects, procedure calls, and the invocation of microservices. Its feasibility is due to the availability of source code in ERTMS/ETCS L2 implementations, as these systems require thorough white-box analyses to certify their safety levels. Fig. \ref{ERTMS_DEPLOYMENT_DIAGRAM} shows an example deployment diagram where the on-board and trackside subsystems communicate with each other and send run-time events to a monitor server. These run-time events configure a partial trace, and are checked against the execution model by a fault detector using a conformance checker that implements process mining techniques.

\subsubsection{Online Conformance Checking} Similarly to process discovery, conformance checking can also be performed through a variety of algorithms, such as footprint-based, token-based, and alignment-based algorithms \cite{aalst2022pmhb}. In this paper, we focus on alignment-based algorithms. These algorithms aim to build an alignment between a trace and the normative model by finding the path across it --- the model trace --- that either matches or is the best approximation of the trace. Using the approach in \cite{vitale2025cfad}, the mismatches between the $k$ actual traces and their corresponding model traces can be translated into ``conformance checking diagnoses", i.e., tabular data $\mathcal{D}\in\mathbb{N}^{k\times m}$, where $m$ is the cardinality of the set of activities $\mathcal{A}$. Each row of $\mathcal{D}$ represents the conformance checking diagnosis of a trace $\sigma$: $d_{\sigma} = \{d_1,\dots,d_{m}\}$, where $d_i$ is the number of misalignments associated with activity $a_i\in\mathcal{A}$.
We aim to perform online conformance checking by employing a window-based approach \cite{burattin2020streamingpdcc}, i.e., once a certain number of run-time events are collected from the system, these compose the partial trace and a partial alignment is computed. The partial trace is enlarged incrementally as more windows of activities are collected.

\subsection{Anomaly Detection and Explanation}
In the third phase, the conformance-checking diagnoses of non-conformant traces are analyzed to spot the occurrence of specific anomalies. As the data is unlabeled, the detection, explanation, labeling, and localization of anomalies are conducted in an unsupervised manner.

\subsubsection{Clustering}
The literature offers a wide variety of algorithms for unsupervised anomaly detection, including, but not limited to \cite{goldstein2016comparative}, \cite{zoppi2021unsupervised}: angle-based, neighbour(density)-based, statistical, neural networks, or clustering. Particularly, clustering is a type of unsupervised machine learning that aims at discovering clusters (i.e., subsets of data points, or diagnoses that are similar between each other) in training data, allowing to map a novel data point, or diagnosis $d_{\sigma}$, to a specific cluster, or no cluster at all (i.e., an outlier). Clustering techniques are particularly fitting for this study, as they partition the input space into ``groups'' that are easy to visualize and easier to understand than with other methods, which are not always transparent to the user. In our methodology, each tuple of diagnoses $d_{\sigma}\in\mathcal{D}$ is assigned to a cluster using one of the techniques above.

\subsubsection{Anomaly Explanation}
\label{anomaly_explanation}
This step aims to explain the different clusters formed with $\mathcal{D}$, analyzing the misalignments of each tuple belonging to the different clusters. Specifically, given $C=\{c_1,c_2,\dots,c_o\}$ the set of $o$ clusters and $\mathcal{D}_c\subseteq\mathcal{D}$ the set of diagnoses belonging to $c\in C$, the anomaly explanation associated with cluster $c$ is vector $s_{c}\in\mathbb{R}^m$ such that $s_c(i)=\sum_{j=1}^{l}d_{j}(i)$, where $i$ is the $i$-th activity $a_i\in\mathcal{A}$. Next, given $COMP=\{comp_1,comp_2,\dots,comp_p\}$ the set of $p$ ERTMS/ETCS L2 components of the target scenario and $comp\in COMP$ a specific component, we compute $s_{c,comp}=\sum_{j\in\mathcal{A}_{comp}} s_c(j)$, which provides the amount of misalignments of the activities $\mathcal{A}_{comp}\subseteq\mathcal{A}$ linked to $comp$. $s_{c,COMP}=\{s_{c,comp_1}, s_{c,comp_2}, \dots,s_{c,comp_p}\}$ is the anomaly explanation of cluster $c$. Finally, we calculate the probabilities of each component $comp$ for a given cluster $c$ by computing $P(c,comp)=\frac{s_{c,comp}}{\sum_{comp_i\in COMP}s_{c,comp_i}}$. These values denote the probabilities that an anomaly occurred in the specific components, and are used in the subsequent labeling step.

\subsubsection{Labeling} \label{labeling} Once the probabilities of each cluster are computed, we are able to label the clusters based on the maximum probability. Specifically, if a cluster contains diagnoses that are more likely to have misalignments in a specific component, the cluster will be labeled as corresponding to an anomaly in such component. Note that this process of deriving explanations and compute the a-posteriori labeling of a clustering result does not require any knowledge of labels (thus it is entirely unsupervised) and solely relies on knowing which steps of the process are related to a specific component, which is something that can be easily derived by any functional specification of a problem, system, or protocol. 

\subsubsection{Anomaly Localization}
When a new trace $\sigma_{test}$ is collected from run-time monitoring, its diagnosis $d_{\sigma_{test}}$ is computed through online alignment-based conformance checking. Based on $d_{\sigma_{test}}$, the trace is assigned to the closest cluster and consequently labeled as a $comp_i$ anomaly if the closest cluster was previously labeled as containing diagnosis tuples affected by a fault in the i-th component. This completes the detection and explanation step, which uses diagnoses (non-conformances) of a trace with respect to its expected process flow to detect anomalies, diagnose their root cause, and localize their origin.

\section{RBC/RBC Handover}
\label{sec:case_study}
In this section, we show the application of the three steps of the methodology shown in Fig. \ref{METHODOLOGY} to our ERTMS/ETCS L2 case study: the RBC/RBC Handover scenario.

\subsection{Behavior Characterization}
In this step, we simulate the RBC/RBC Handover scenario through a high-level description of its process according to the SUBSET-026 document of the ERTMS standard. First, we compiled the various sequences identified in the document under the chosen scenario into a BPMN model. Since the model describes a workflow from start to end, the model can be simulated, and various traces can be collected. 

After the BPMN traces are collected, they are further pre-processed to mimic a realistic monitoring process of the computer-based system. Specifically, we aim to mimic the procedural programming paradigm, which is commonly found in the real implementation of ERTMS/ETCS systems. Using the activities of the traces generated through the BPMN model, our approach records the invocation of a procedure each time a new activity is executed as follows. 

Let $X$ be an ERTMS/ETCS component and $\mathcal{P}_X$ be the procedures implemented by $X$. One of the procedures is executed with a given probability for each distinct activity carried out by the component. It is expected that normal behavior follows a given control-flow with a high probability, whereas anomalous behavior involves the alteration of such normal control flow with the improper execution of other procedures. This is enforced using the following probability scheme. Let $\mathcal{A}_{HO}$ be the set of activities of the RBC/RBC Handover scenario, $\mathcal{X}=\{ARBC,HRBC,EVC,RTM\}$ the set of ERTMS/ETCS components involved in the use case, $\mathcal{P}_{a,X}=\{p_0, \dots, p_{q_X}\}$ the $q_X\in\mathbb{N}$ procedures implemented by $X\in\mathcal{X}$, and $\rho\in[0,1)$ a real number. We define a procedure execution probability scheme $PEP$ for each $\mathcal{P}_{a,X}$ associated with $a\in\mathcal{A}_{HO}$ that assigns probability values to the procedures so that one procedure is much more likely to be chosen than others, allowing the simulation of traces that are very similar to each other. In particular, let us consider a real number $\rho\in\mathbb{R}$ and the following function definition:
\begin{equation*}
PEP: \mathbb{R}\times (\mathcal{P}_{a,X})^{q_x}\rightarrow \mathbb{R}^{q_x},
\end{equation*}
$PEP(\rho,p_0,\dots,p_{q_X})=(pr_{p_0},pr_{p_1},\dots,pr_{p_{q_X}})$, where $pr_{p_0}=\rho$ and the other probabilities are equal to $\frac{1-\rho}{q_X-1}$. For example, given $\mathcal{P}_{a,X}=\{p_0, \dots, p_{q_X}\}$, if $\rho=0.99$, the application of $PEP$ to $\mathcal{P}_{a,X}$ is such that $p_0$ is selected with 99\% probability, whereas there is a 0.1\% probability that other procedures are executed. 

Following the procedure above, we generate the execution traces of RBC/RBC Handover as follows.
\begin{enumerate}
    \item Set $q_X$ for each component $X$ involved in the RBC/RBC Handover use case;
    \item For each activity $a\in\mathcal{A}_{HO}$, enforce a random order to the procedures that may be executed for that activity, i.e., generate $\mathcal{P}_{a,X}$;
    \item Set $\rho=0.99$;
    \item Simulate $N_{norm}\in\mathbb{N}$ traces\footnote{Popular simulation tools are \texttt{pm4py} (\url{https://processintelligence.solutions/pm4py}) and \texttt{ProM} (\url{https://promtools.org/}).};
    \item For each activity of the $N_{norm}$ traces, sample a procedure through $PEP$ using $\rho$.
\end{enumerate}
In this preliminary setup, we have decided to set $q_X=10$ for each component and generate $N_{norm}=100$ traces. The dataset generation procedure has been published on the GitHub repository\footnote{\url{https://github.com/francescovitale/pm\_ertms}}.

\begin{figure*}[!t]
\centering
\includegraphics[width=\textwidth]{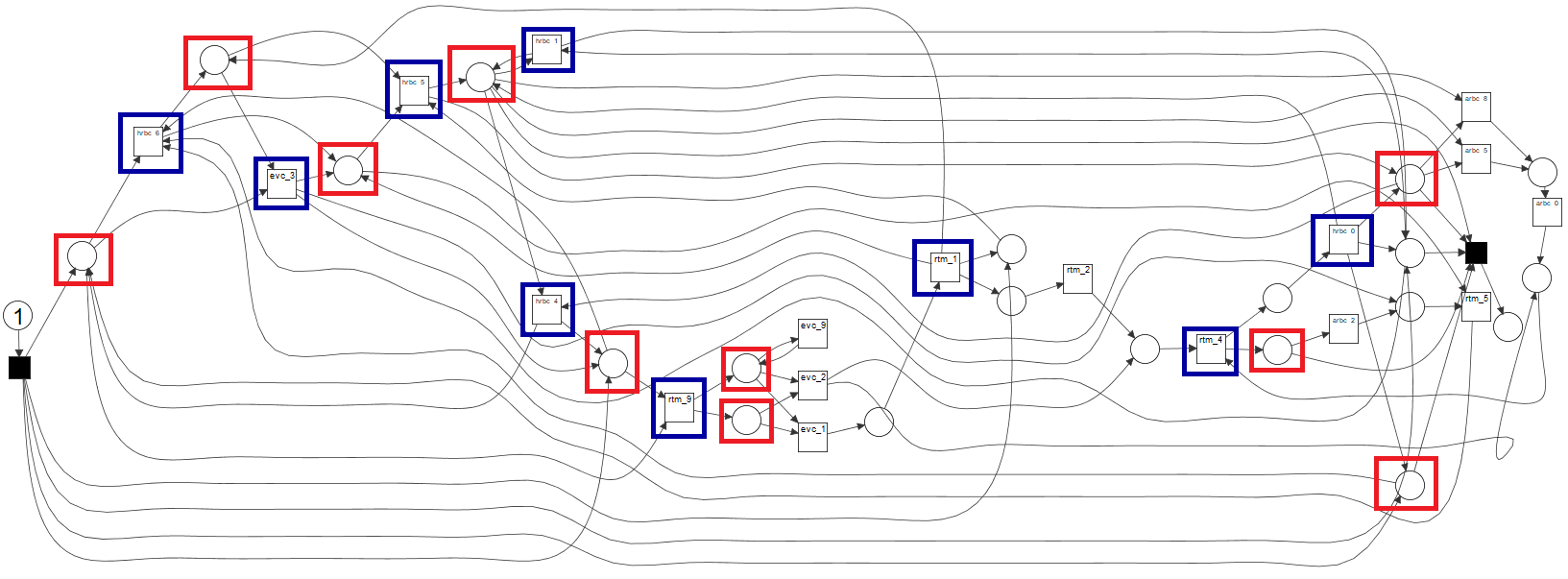}
\caption{The RBC/RBC Handover Petri net obtained through the integer linear programming-based miner during behavior characterization.}
\label{RBC_RBC_HO_PETRI_NET}
\end{figure*}

We used the integer linear programming-based miner with 75\% tolerance to noise to build a workflow Petri net from the $N_{norm}$ traces. Fig. \ref{RBC_RBC_HO_PETRI_NET} shows the Petri net that the algorithm extracted. The Petri net contains two specific places: the source and the sink. The source place, marked with `1', allows triggering the first transition, whereas the sink place marks the end of the workflow. The red boxes outline those places with more than one outgoing arc, which means that in the presence of a single token in the place, only one of the outgoing paths can be followed --- an exclusive control-flow pattern. The blue boxes outline those transitions with more than one outgoing arc, which means that multiple concurrent paths are activated when the transition is triggered --- a concurrent control-flow pattern. The Petri net has 21 places, 20 transitions and 78 arcs. The Woflan diagnosis \cite{verbeek2001woflan} reports that there always exists a path leading to the sink and that there are no dead transitions.

\subsection{Behavior Diagnosis} 
\label{window_size}
In this part, we aim to extract online actions from run-time traces and check them against the execution model obtained in the previous step. We can simulate run-time traces with the same procedure above. In particular, we simulated $N_{anom}=400$ anomalous traces and injected control-flow anomalies. As there are 4 components, we split the $N_{anom}$ traces into four sets of 100 traces, which we refer to as $N_{anom,ARBC}$ traces, $N_{anom,HRBC}$ traces, $N_{anom,EVC}$ traces, and $N_{anom,RTM}$ traces. Each set of traces is injected with a different type of anomaly. The injection procedure is as follows. Let $X$ be one of the four components and let us consider its corresponding set of $N_{anom,X}$ traces. We scan the traces procedure-by-procedure and, if the procedure is executed by $X$, we randomly inject either a wrongly-ordered, skipped or wrong procedure control-flow anomaly. To account for the randomness of the simulation, we replicated the simulation ten times with different random seeds.

After we obtained the four sets of traces, we proceeded with the extraction of online actions by splitting each trace into multiple traces according to a window size. The number of online actions during this phase depends on the window size. We set three different window sizes, 5, 10 and 15, to evaluate the change in the results of the subsequent steps. For example, if a given trace has 20 traces and the window size is 10, it is split into 2 subtraces, where the first subtrace contains the first 10 online actions, whereas the second subtrace is the entire trace. Regardless of the window size, each subtrace is checked against the execution model to collect the conformance checking diagnoses to use in the subsequent steps.

\subsection{Anomaly Detection and Explanation}
The conformance checking diagnoses of the four sets of traces are clustered to identify the different anomalies. In particular, to evaluate the quality of different clustering techniques in correctly classifying the anomalies, we split the diagnoses into a training and a test set. We held out 25\% of the diagnoses and used the training set to extract the clusters. Hence, there are 300 training diagnoses and 100 test diagnoses. Each cluster obtained from the training set was labeled with the type of anomaly injected. For example, all the clusters formed with the diagnoses of $N_{anom,ARBC}$ traces are labeled ARBC. The test diagnoses are subsequently assigned to each cluster and labeled accordingly. For completeness of our analysis, we selected multiple clustering algorithms, ensuring to select candidate clustering algorithms that are as diverse as possible between each other. We selected clustering algorithms that rely (e.g., K-Means) or do not rely on centroids (e.g., DBSCAN), algorithms that automatically derive the optimal number of clusters, and some that require the user to input the desired number of clusters as a parameter. Overall, we selected 10 algorithms: BIRCH, K-Means, Mini-Batch KMeans, Ward Hierarchical Clustering, Spectral Clustering, HDBSCAN, DBSCAN, OPTICS, MeanShift, AffinityPropagation, whose implementation is available in the open-source Python library \texttt{scikit-learn}. Algorithms that required the number of clusters as input were exercised 5 times, each with the input parameter of 10, 30, and 50 clusters. 

The conformance checking diagnoses collect the misalignments of the 40 possible procedures. These procedures are categorized according to the component that executes them: procedures 0 to 9: ARBC component; procedures 10 to 19: EVC component; procedures 20 to 29: HRBC component; procedures 30 to 39: RTM component, thus $comp = \{ARBC, EVC, HRBC, RTM\}$. For each cluster, the anomaly explanation $s_c$ will be used to compute the quantities $s_{c, ARBC}$, $s_{c, EVC}$, $s_{c, HRBC}$, $s_{c, RTM}$, each quantifying the per-activity misalignment. Consequently, and regardless of the specific clustering algorithm, each cluster is labeled according to the process in Section \ref{labeling} either as ARBC, EVC, HRBC or RTM. 

\section{Experimentation}
\label{sec:experimentation}
Our experiments aim at evaluating the ability of the methodology to localize the different faults in the presence of control-flow anomalies.

\subsection{Evaluation Metrics}
There is a wide variety of metrics to measure the classification performance of multi-class classifiers \cite{hossin2015review}, the most common being \textit{accuracy}, typically used in its \textit{balanced} formulation. Other commonly used metrics are those that are primarily meant for binary classification (e.g., precision, recall, F-measure, matthews correlation coefficient) but may be adapted to multi-class scenarios by performing an average per class. For this study, we choose the \textit{balanced accuracy} to quantify the performance of clustering algorithms.

In addition to the classification metrics, there is a subset of metrics that aim at assessing the performance of clustering processes. These metrics stem from the idea that a cluster should contain all and only instances of a specific class. To evaluate such capability, two metrics are usually employed: homogeneity, which determines whether a cluster contains coherent data points, and completeness, which evaluates whether data points of a given class are elements of the same cluster. The V-measure \cite{rosenberg2007v} computes the harmonic mean of homogeneity and completeness. In this study, we use the V-measure to evaluate the quality of the clustering process.

Finally, we evaluate the effectiveness of the anomaly explanation mechanism discussed in Section \ref{anomaly_explanation}. However, in this case, instead of computing the per-cluster explanation, we consider the explanation of each trace in the test set individually. Then, given a specific component, we sum all the test trace explanations, obtaining the four global anomaly explanations $S_{ARBC}$, $S_{EVC}$, $S_{HRBC}$ and $S_{RTM}$.

\subsection{Anomaly Detection Results}

\begin{table}[!t]
\centering
\caption{Anomaly detection results (best values are bolded in the table).}
\label{CLUSTERING_RESULTS}
\resizebox{\textwidth}{!}{%
\begin{tabular}{clllllllllllllllllll}
\toprule
\multirow{2}{*}{\textbf{\begin{tabular}[c]{@{}c@{}}Window\\ size\end{tabular}}} & \textbf{}       &  & \multicolumn{3}{l}{\textbf{K-Means}} &  & \multicolumn{3}{l}{\textbf{WARD}} &  & \multicolumn{3}{l}{\textbf{Spectral}} &  & \multicolumn{3}{l}{\textbf{Birch}} &  & \multicolumn{1}{c}{\textbf{HDBScan}} \\ \cline{4-6} \cline{8-10} \cline{12-14} \cline{16-18} \cline{20-20} 
                                                                                  & \textbf{Metric} &  & 10         & 30         & 50         &  & 10        & 30        & 50        &  & 10       & 30      & 50               &  & 10         & 30        & 50        &  & -                                    \\ \hline
\multirow{2}{*}{5}                                                                & Accuracy        &  & 0.61       & 0.78       & 0.82       &  & 0.68      & 0.80      & 0.83      &  & 0.61     & 0.80    & 0.83             &  & 0.62       & 0.74      & 0.76      &  & 0.69                                 \\
                                                                                  & V-measure       &  & 0.32       & 0.52       & 0.56       &  & 0.42      & 0.56      & 0.59      &  & 0.32     & 0.53    & 0.57             &  & 0.33       & 0.48      & 0.50      &  & 0.37                                 \\
\multirow{2}{*}{10}                                                               & Accuracy        &  & 0.75       & 0.86       & 0.88       &  & 0.70      & 0.87      & 0.89      &  & 0.77     & 0.87    & 0.89             &  & 0.72       & 0.85      & 0.86      &  & 0.61                                 \\
                                                                                  & V-measure       &  & 0.52       & 0.64       & 0.68       &  & 0.47      & 0.69      & 0.72      &  & 0.54     & 0.66    & 0.70             &  & 0.50       & 0.67      & 0.66      &  & 0.35                                 \\
\multirow{2}{*}{15}                                                               & Accuracy        &  & 0.81       & 0.89       & 0.91       &  & 0.75      & 0.88      & 0.89      &  & 0.82     & 0.92    & \textbf{0.94}    &  & 0.77       & 0.88      & 0.87      &  & 0.34                                 \\
                                                                                  & V-measure       &  & 0.63       & 0.73       & 0.75       &  & 0.57      & 0.71      & 0.72      &  & 0.61     & 0.77    & \textbf{0.81}    &  & 0.58       & 0.73      & 0.70      &  & 0.09                                 \\ \hline
\end{tabular}%
}
\end{table}
\begin{figure*}[!t]
\centering
\begin{subfigure}{0.48\textwidth}
    \includegraphics[width=\linewidth]{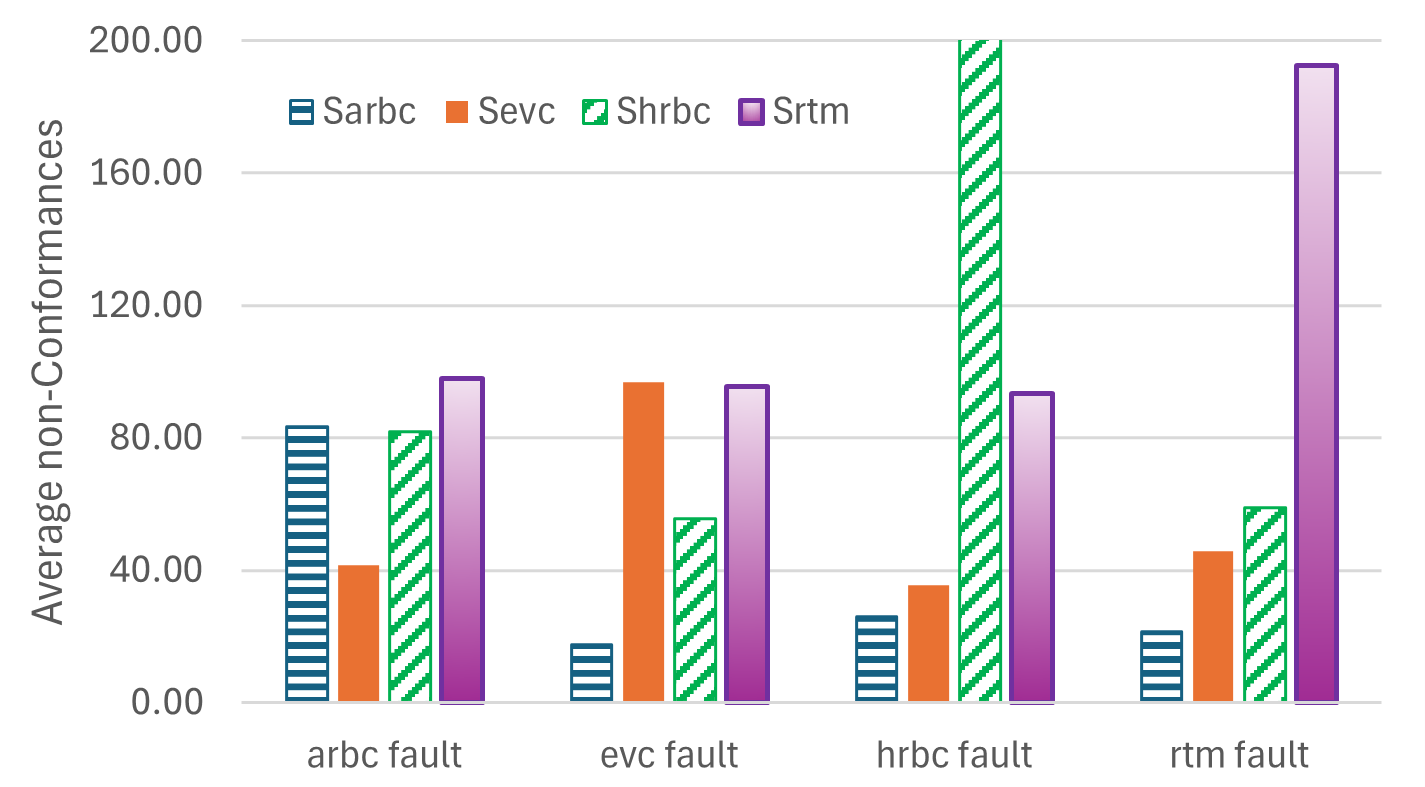}
    \caption{Window size 5}
    \label{clustering_plots_1}
\end{subfigure}
\hfill
\begin{subfigure}{0.48\textwidth}
    \includegraphics[width=\linewidth]{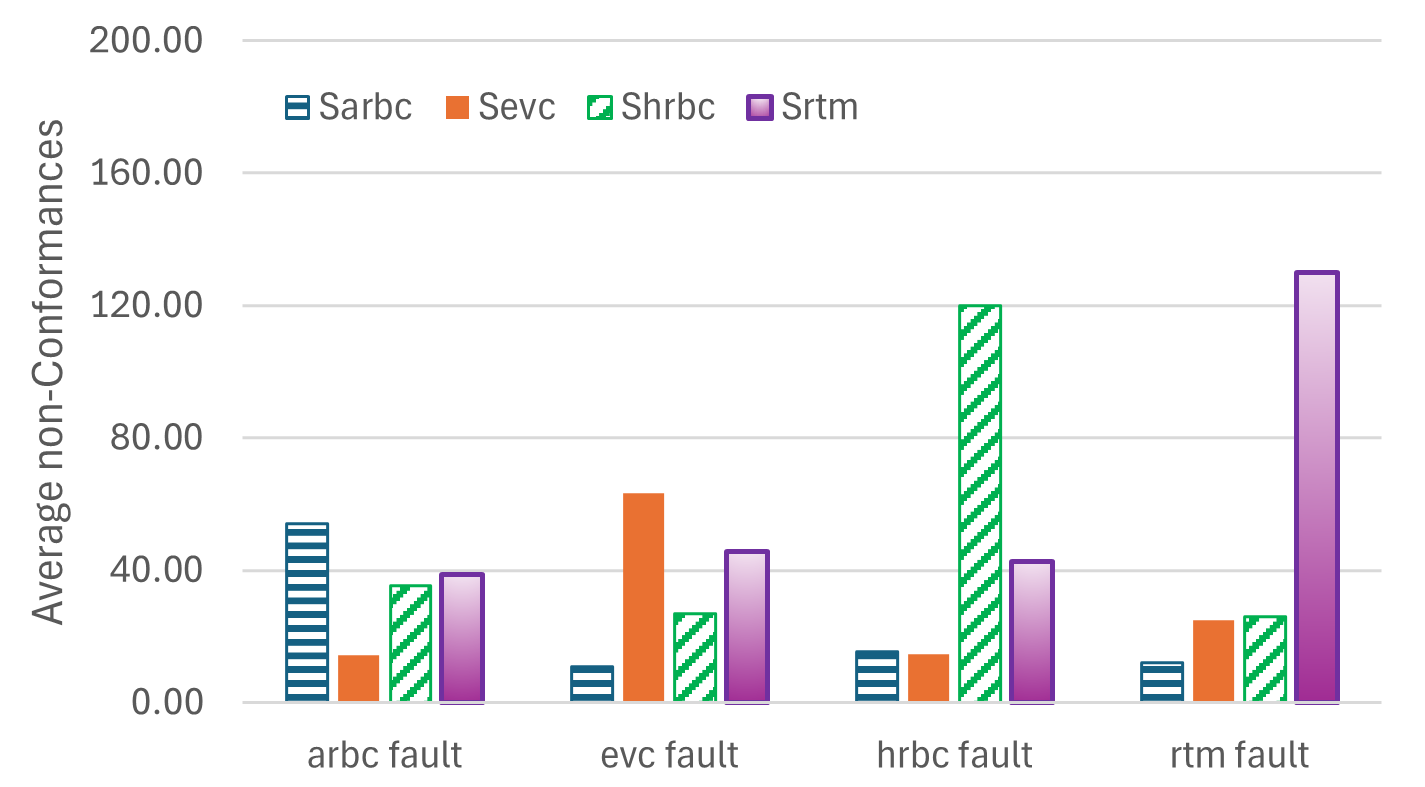}
    \caption{Window size 10}
    \label{clustering_plots_2}
\end{subfigure}

\vspace{0.5em}

\begin{subfigure}{0.48\textwidth}
    \centering
    \includegraphics[width=\linewidth]{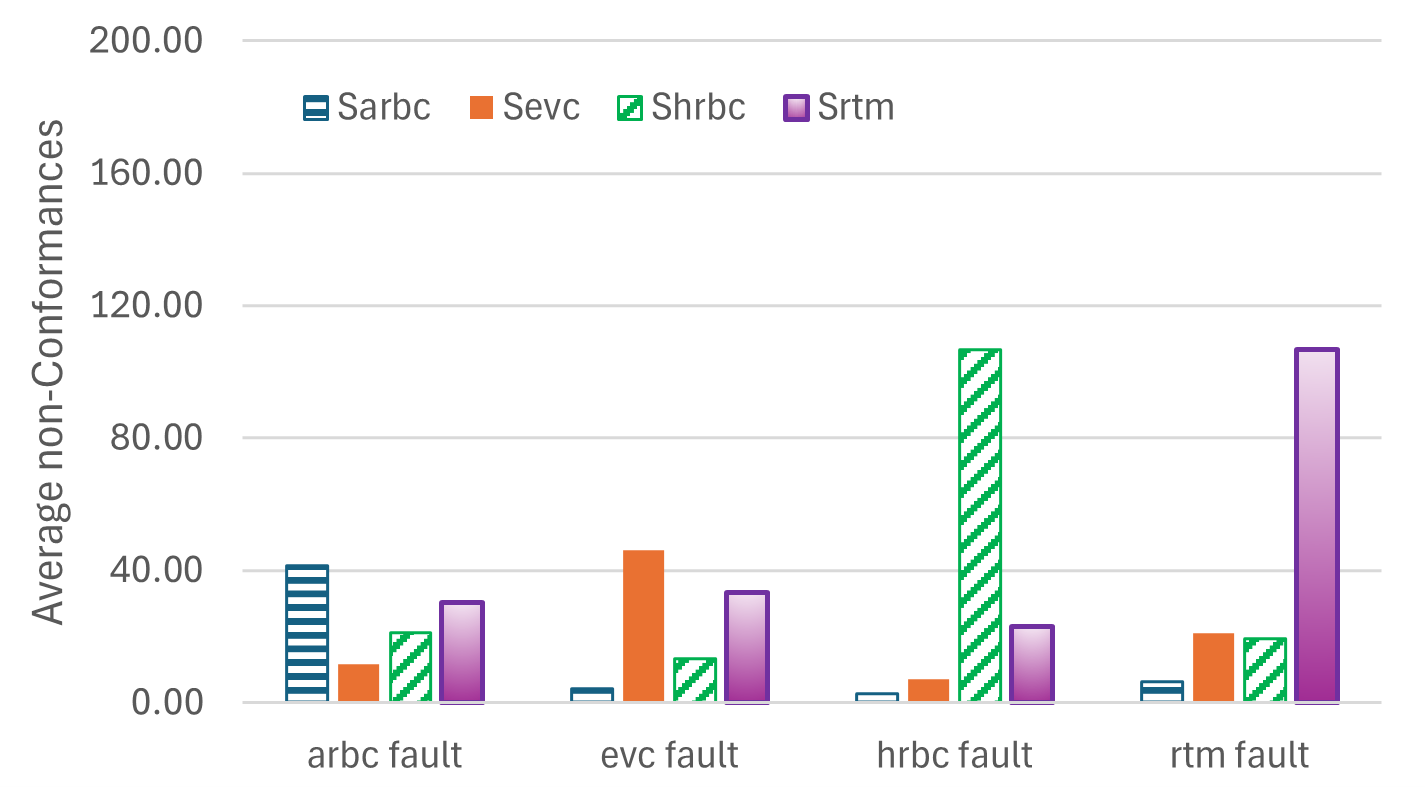}
    \caption{Window size 15}
    \label{clustering_plots_3}
\end{subfigure}

\caption{Spectral clustering with 50 clusters, averaged over experiments with different injected faults.}
\label{clustering_plots}
\end{figure*}

Table \ref{CLUSTERING_RESULTS} reports the anomaly detection results in terms of accuracy and V-measure for different window sizes and clustering algorithms. We limit our analysis to the best-performing clustering algorithms requiring the manual tuning of the number of clusters (K-Means, WARD, Spectral Clustering, and BIRCH) and the best-performing density-based clustering algorithm (HDBScan).

First, let us consider the impact of increasing the number of clusters in K-Means, WARD, Spectral Clustering and BIRCH on the evaluation metrics. Except for BIRCH, increasing the number of clusters leads to better performance, with Spectral Clustering peaking at 0.94 accuracy and 0.81 V-measure with 50 clusters. Such high performance is due to the ability of the algorithm to build specific, homogeneous, and complete clusters. However, it is worth noting that a higher cluster number increases the computational complexity and may lead to overfitting due to the specificity of the clusters being built. Therefore, selecting a lower number of clusters could be safer in terms of generalization capability and speed while maintaining a reasonably high accuracy, as Spectral Clustering achieves 0.82 accuracy and 0.61 V-measure with only 10 clusters.

Second, let us discuss the impact of the window size on the evaluation metrics. On the one hand, a window size equal to 5 leads to worse performance, with the lowest accuracy and V-measure being 0.61 and 0.32 for K-Means. On the other hand, increasing the window size leads to higher accuracy and V-measure across all the clustering algorithms --- except for HDBScan, whose accuracy and V-measure, respectively, drop from 0.69 to 0.34 for window size equal to 5 and 0.37 to 0.09 for window size equal to 15. Hence, while K-Means, WARD, Spectral Clustering, and BIRCH are able to build better clusters through the additional conformance checking diagnoses yielded by a larger window size, HDBScan is unable to effectively use such information. In conclusion, it is worth noting that, despite the majority of algorithms achieving better accuracy and V-measure, increasing the window size generally leads to higher conformance checking times.

In conclusion, online conformance checking of the online traces with the Petri net in Fig. \ref{RBC_RBC_HO_PETRI_NET} allowed us to extract diagnoses that can be accurately handled by our clustering process to predict the fault that occurred in the system, achieving up to 0.94 accuracy and 0.81 V-measure.

\subsection{Anomaly Explanation Results}
The bar plots of Figs. \ref{clustering_plots_1}, \ref{clustering_plots_2}, and \ref{clustering_plots_3} show the anomaly explanations $S_{ARBC}$, $S_{EVC}$, $S_{HRBC}$ and $S_{RTM}$ for window sizes 5, 10, and 15, respectively. These results are based on the clusters built with Spectral Clustering using 50 clusters (the best clustering strategy from the previous discussion). $S_{ARBC}$, $S_{EVC}$, $S_{HRBC}$, and $S_{RTM}$ are divided into four batches. Each batch collects the anomaly explanations related to the test traces of a specific fault type. For example, the first batch on the left-hand side of each bar plot collects the anomaly explanations of the $N_{anom,ARBC}$ traces.

Although it is expected that the activation of a fault within a specific component $comp$ leads to high $S_{comp}$ with a clear separation from the other explanations, some fault type-window size configurations fail to achieve this result. For example, the results of a window size equal to 5 shown in Fig. \ref{clustering_plots_1} show that $S_{ARBC}$ is lower than $S_{RTM}$ despite the activation of an ARBC fault. However, this does not occur for window sizes equal to 10 (Fig. \ref{clustering_plots_2}) and 15 (Fig. \ref{clustering_plots_3}), which is consistent with the accuracy and V-measure reported in Table \ref{CLUSTERING_RESULTS}. Still, while window sizes equal to 10 and 15 allow better localization of the ARBC and EVC faults, the separation of the different anomaly explanations is not so clear. On the other hand, the HRBC and RTM faults always lead to, respectively, high $S_{HRBC}$ and $S_{RTM}$, and, in both cases, a very clear separation of anomaly explanations regardless of the window size. This suggests that faults in specific components are more likely to propagate to other components, making the labeling and localization of the fault more challenging in any window size scenario.

In summary, our explanation process allowed us to identify the misbehaving components of different online traces. The larger the window size, the more accurate the localization is with respect to the specific faulty component.

\section{Conclusions}
\label{sec:conclusions}

In this paper, we presented a novel approach for run-time control-flow anomaly detection in ERTMS/ETCS L2 using process mining. By leveraging execution traces and conformance checking, the behavior of the system can be monitored to identify deviations from expected execution models. Our method also integrates unsupervised machine learning to support anomaly detection and localization, offering insights into the root causes of observed deviations with a high degree of accuracy and explainability. The RBC/RBC Handover scenario has been chosen as a good-fitting use case of complex coordination tasks in train control; as such, it served as a reference case-study to validate our approach through a proof-of-concept, demonstrating its feasibility and effectiveness in realistic settings.

This work contributes to enhancing the dependability and fault-tolerance of safety-critical railway systems by bridging the gap between offline verification and online assurance. In line with the growing need for more resilient and intelligent railway software, our future research will focus on incorporating self-adaptive capabilities, wherein detected anomalies may trigger autonomous mitigation actions or adaptation strategies. Furthermore, the integration of our monitoring framework within digital twin architectures of ERTMS/ETCS subsystems is foreseen, enabling predictive diagnostics and closed-loop assurance \cite{debenedictis2023dtadiiot, vitale2025pmdt}.

Ultimately, our goal is to contribute to a new generation of intelligent, trustworthy railway systems capable of continuous assurance in the face of evolving operational contexts and threats.

\begin{credits}
\subsubsection{\ackname} The work of Francesco Vitale and Nicola Mazzocca was partly supported by the Spoke 9 “Digital Society \& Smart Cities” of ICSC - Centro Nazionale di Ricerca in High Performance-Computing, Big Data and Quantum Computing, funded by the European Union - NextGenerationEU (PNRR-HPC, CUP: E63C22000980007). 
The work of Francesco Flammini was partly supported by the Swiss State Secretariat for Education, Research and Innovation (SERI) under contracts no. 23.00321 (Academics4Rail) and 24.00528 (PhDs EU-Rail); those projects have been selected within the European Union’s Horizon Europe research and innovation programme under grant agreements no. 101121842 and 101175856, respectively.
The work of Tommaso Zoppi was partly supported by the project SERICS (PE00000014) under the MUR National Recovery and Resilience Plan funded by the European Union – NextGenerationEU, and by the Cognitive Safety with Point Clouds (CogniSafe3D) Eurostars 3, Call 6 by the European Union. Views and opinions expressed are however those of the authors only and do not necessarily reflect those of the funding agencies, which cannot be held responsible for them.

\subsubsection{\discintname}
The authors have no competing interests to declare that are
relevant to the content of this article.
\end{credits}

\bibliographystyle{splncs04}
\bibliography{bibliography}

\end{document}